\documentclass[journal]{IEEEtran}
\ifCLASSINFOpdf
\else
\fi

\usepackage{times}
\usepackage{epsfig}
\usepackage{graphicx}
\usepackage{amsmath}
\usepackage{dsfont}
\usepackage{amssymb}
\usepackage{url}
\usepackage{amsfonts}
\usepackage{multirow}
\usepackage{makecell}
\usepackage{booktabs}
\usepackage{lineno}
\usepackage{caption}
\usepackage{subcaption}

\begin{document}

\title{Multimodal Transformer with Multi-View\\Visual Representation for Image Captioning}
\author{Jun~Yu, \IEEEmembership{Member, IEEE},~
        Jing~Li,~
        Zhou~Yu, \IEEEmembership{Member, IEEE},~
        Qingming~Huang, \IEEEmembership{Fellow, IEEE}~
\thanks{J. Yu, J. Li and Z. Yu are with Key Laboratory of Complex Systems Modeling and Simulation,
School of Computer Science and Technology, Hangzhou Dianzi University, Hangzhou, 310018, China (e-mail: yujun@hdu.edu.cn; jingli@hdu.edu.cn; yuz@hdu.edu.cn).}
\thanks{Q. Huang is with the School of Computer and Control Engineering, University of Chinese Academy of Sciences, Beijing 101408, China (email: qmhuang@ucas.ac.cn).}
\thanks{Corresponding author: Zhou Yu.}
        }

\markboth{Journal of \LaTeX\ Class Files,~Vol.~14, No.~8, August~2015}%
{Yu \MakeLowercase{\textit{et al.}}: Multimodal Transformer with Multi-View Visual Representation for Image Captioning}
%



\maketitle


\begin{abstract}
Image captioning aims to automatically generate a natural language description of a given image, and most state-of-the-art models have adopted an encoder-decoder framework. The framework consists of a convolution neural network (CNN)-based image encoder that extracts region-based visual features from the input image, and an recurrent neural network (RNN)-based caption decoder that generates the output caption words based on the visual features with the attention mechanism. Despite the success of existing studies, current methods only model the co-attention that characterizes the inter-modal interactions while neglecting the self-attention that characterizes the intra-modal interactions.
Inspired by the success of the Transformer model in machine translation, here we extend it to a Multimodal Transformer (MT) model for image captioning. Compared to existing image captioning approaches, the MT model simultaneously captures intra- and inter-modal interactions in a unified attention block. Due to the in-depth modular composition of such attention blocks, the MT model can perform complex multimodal reasoning and output accurate captions. Moreover, to further improve the image captioning performance, multi-view visual features are seamlessly introduced into the MT model. We quantitatively and qualitatively evaluate our approach using the benchmark MSCOCO image captioning dataset and conduct extensive ablation studies to investigate the reasons behind its effectiveness. The experimental results show that our method significantly outperforms the previous state-of-the-art methods. With an ensemble of seven models, our solution ranks the 1st place on the real-time leaderboard of the MSCOCO image captioning challenge at the time of the writing of this paper.
\end{abstract}

\begin{IEEEkeywords}
Image captioning, multi-view learning, deep learning.
\end{IEEEkeywords}

\section{Introduction}
\IEEEPARstart{R}ecent advances in deep learning have resulted in great progress in both the computer vision and natural language processing communities. These achievements make it possible to connect vision and language, and facilitate multimodal learning tasks such as image-text matching \cite{yu2014discriminative}, visual question answering \cite{fukui2016multimodal}\cite{kim2018bilinear}\cite{yu2017mfb}, visual grounding \cite{yu2018rethinking} and image captioning \cite{anderson2018bottom}\cite{xu2016ask}\cite{xu2018dual}\cite{lu2017knowing}\cite{rennie2017self}.

Image captioning aims to automatically describe an image's content using a natural language sentence. The task is challenging since it requires one to recognize key objects in an image, and to understand their relationships with each other. Most successful image captioning approaches adopt the encoder-decoder framework, which is inspired by the sequence-to-sequence model for machine translation \cite{sutskever2014sequence}. The framework consists of a convolutional neural network (CNN)-based image encoder that extracts region-based visual features from an input image, and an recurrent neural network (RNN)-based caption decoder that iteratively generates the output caption words based on the visual features. The encoder-decoder model is usually trained in an end-to-end manner to minimize the cross-entropy loss. Based on the framework, plenty of improvements have been made by recent works to further improve image captioning performance further. For instance, to establish the fine-grained connections of caption words and their related image regions, an attention mechanism can be seamlessly inserted into the framework \cite{xu2016ask}. To provide a better understanding of the objects in the image, region-based bottom-up-attention features can be extracted from a pre-trained object detector to replace the traditional CNN convolutional features \cite{anderson2018bottom}. To address the exposure bias of generated captions by using the cross-entropy loss, reinforcement learning (RL)-based algorithms are designed to directly optimize the non-differentiable evaluation metrics (\emph{e.g.,} BLEU \cite{papineni2002bleu} and CIDEr \cite{vedantam2015cider}) \cite{rennie2017self}.

Despite the success that existing approaches have achieved, they have the following limitations: 1) the current attention mechanism in image captioning only models the \emph{co-attention} that characterizes inter-modal interactions (\emph{i.e.}, object-to-word) while neglecting the \emph{self-attention} that characterizes intra-modal interactions (\emph{i.e.}, word-to-word and object-to-object); 2) current image captioning models are usually \emph{shallow} and may fail to fully understand the complex relationships among visual objects; and 3) the region-based visual features may fail to cover all objects in the image, leading to insufficient visual representations for generating accurate captions.

To address the first and second limitations, we extend the Transformer model for machine translation \cite{vaswani2017attention} to a Multimodal Transformer (MT) model for image captioning. Different from the CNN-RNN captioning models, the MT model does not use RNN and instead relies entirely on an attention mechanism to assess the global dependencies between the input and output. By properly stacking such attention blocks in depth, MT forms a deep encoder-decoder model that simultaneously captures the self-attention within each modality and the co-attention across different modalities. To address the last limitation, we introduce multi-view feature learning into the MT model to adapt both the aligned and unaligned multi-view visual features.

To summarize, the main contributions of this study are three-fold:
\begin{itemize}
\item The joint modeling of the self-attention and the co-attention interactions for image captioning is first proposed in the MT model. The MT model is capable of modeling three types of relationships using a modular attention block, \emph{i.e.,} word-to-word, object-to-object, and word-to-object. By stacking such attention blocks in depth, the deep MT model significantly outperforms the state-of-the-art models, thereby highlighting the importance of deep reasoning for image captioning.
\item Multi-view learning on the image is introduced in conjunction with the MT model to provide more diverse and discriminative visual representations. We introduce two alternative strategies to handle aligned and unaligned multi-view features, respectively.
\item Extensive experiments on the benchmark MSCOCO image captioning dataset are conducted to quantitatively and qualitatively prove the effectiveness of the proposed models. The experimental results show that the MT significantly outperforms previous state-of-the-art approaches with a single model. Furthermore, our solution ranks the 1st place on the real-time leaderboard of the MSCOCO image captioning challenge with an ensemble of MT models.
\end{itemize}

The rest of the paper is organized as follows: In section \ref{sec:related_work}, we review the related work of image captioning approaches, especially the ones introducing attention mechanisms. In section \ref{sec:mt}, we revisit the basic Transformer model and then propose the Multimodal Transformer model for image captioning. In section \ref{sec:multiview}, we introduce multi-view image representation into the MT model to increase the visual representation capacity, and the quality of the generated captions. In section \ref{sec:exp}, we introduce our extensive experimental results for algorithm evaluation and use the benchmark MSCOCO image captioning dataset to evaluate our proposed approaches. Finally, we conclude this work in section \ref{sec:conclusion}.

\section{Related Work}\label{sec:related_work}
In this section, we briefly review the most relevant research on image captioning, especially those studies that introduce attention models.

\subsection{Image Captioning}
The research on image captioning can be categorized into the following three classes: template-based approaches \cite{kulkarni2013babytalk}\cite{mitchell2012midge}\cite{yang2011corpus}, retrieval-based approaches \cite{karpathy2014deep}\cite{farhadi2010every}\cite{devlin2015language}, and generation-based approaches \cite{rennie2017self}\cite{lu2017knowing}\cite{yao2017boosting}\cite{anderson2018bottom}\cite{yao2018exploring}.

The template-based approaches address the task using a two-stage strategy: 1) align the sentence fragments (\emph{e.g.}, subject, object, and verb) with the predicted labels from the image; and 2) generate the sentence from the segments using pre-defined language templates. Kulkarni \emph{et al.} use the conditional random field (CRF) model to predict labels based on the detected objects, attributes, and prepositions, and then generate caption sentences with a template by filling in the blanks with the most likely labels \cite{kulkarni2013babytalk}. Yang \emph{et al.} employ the HMM model to select the best objects, verbs, and prepositions with respect to the log-likelihood for segments generation \cite{yang2011corpus}. Intuitively, the captions that are generated by the template-based approaches highly depend on the quality of the templates and usually follow the syntactical structures. However, the diversity of the generated captions is severely restricted.

To ease the diversity problem, retrieval-based approaches are proposed to \emph{search} the most relevant captions from a large-scale caption database with respect to their cross-modal similarities to the given image. Karpathy \emph{et al.} propose a deep fragment embedding approach to match the image-caption pairs based on the alignment of visual segments (the detected objects) and caption segments (subjects, objects, and verbs) \cite{karpathy2014deep}. In the testing stage, the cross-modal matching over the whole caption database (usually the captions from the training set) is performed to generate the caption for one image. Other methods such as \cite{farhadi2010every}\cite{devlin2015language} use different metrics or loss functions to learn the cross-modal matching model. However, the retrieval efficiency becomes a bottleneck for these approaches when the caption database is large and restricting the size of the database may reduce the caption diversity. Moreover, retrieval-based approaches cannot generate novel captions beyond the database, which means the diversity problem has not been completely resolved.

Different from template-based and retrieval-based models, generation-based models aim to learn a language model that can generate novel captions with more flexible syntactical structures. With this purpose, recent works explore this direction by introducing the neural networks for image captioning. Vinyals \emph{et al.} propose an encoder-decoder architecture by utilizing the GoogLeNet \cite{szegedy2015going} and LSTM networks \cite{hochreiter1997long} as its backbones. Similar architectures are also proposed by Donahue \emph{et al.} \cite{donahue2015long} and Karpathy \emph{et al.} \cite{karpathy2015deep}. Due to the flexibility and excellent performance, generation-based models have become the mainstream for image captioning.

\subsection{Attention Mechanism}
Within the encoder-decoder framework, one of the most important improvements for generation-based models is the attention mechanism. Xu \emph{et al.} introduce the soft and hard attention models to mimic the human eye focusing on different regions in an image when generating different caption words. The attention model is a \emph{pluggable} module that can be seamlessly inserted into previous approaches to remarkably improve the caption quality. The attention model is further improved in \cite{anderson2018bottom}\cite{chen2017sca}\cite{lu2017knowing}\cite{rennie2017self}. Anderson \emph{et al.} introduce a bottom-up module, that uses a pre-trained object detector to extract region-based image features, and a top-down module that utilizes soft attention to dynamically attend to these object \cite{anderson2018bottom}. Chen \emph{et al.} propose a spatial- and channel-wise attention model to attend to visual features \cite{chen2017sca}. Lu \emph{et al.} present an adaptive attention encoder-decoder model for automatically deciding when to rely on visual or language signals \cite{lu2017knowing}. Rennie \emph{et al.} design a FC model and an Att2in model that achieve good performance \cite{rennie2017self}.

Beyond the image captioning tasks, attention mechanisms are widely used in other multi-modal learning tasks such as visual question answering (VQA). Lu \emph{et al.} propose a co-attention learning framework to alternately learn the image attention and question attention \cite{lu2016hierarchical}. Yu \emph{et al.} reduce the co-attention method into two steps, self-attention for a question embedding and the question-conditioned attention for a visual embedding \cite{yu2018beyond}. Nam \emph{et al.} propose a multi-stage co-attention learning model to refine the attentions based on the memory of previous attentions \cite{nam2016dual}. However, these co-attention models learn separate attention distributions for each modality (image or question) and neglect the dense interaction between each question word and each image region, which becomes a bottleneck for understanding the fine-grained relationships of multimodal features. To address this issue, dense co-attention models have been proposed, which establish the complete interaction between each question word and each image region \cite{nguyen2018improved}\cite{kim2018bilinear}. Compared to the previous co-attention models with coarse interactions, the dense co-attention models deliver significantly better VQA performance.

\section{Multimodal Transformer}\label{sec:mt}
In this section, we first briefly describe the preliminary knowledge of the Transformer model \cite{vaswani2017attention}. Then, we introduce the proposed Multimodal Transformer (MT) framework for image captioning, which consists of an \emph{image encoder} and a \emph{caption decoder}. The image encoder learns the deep image representation in a self-attention manner, and then, the caption decoder uses the attended image representations to generate textual captions.

\subsection{The Transformer Model}
The Transformer model \cite{vaswani2017attention} was first proposed for machine translation, and has been successfully applied to many natural language processing tasks. We first introduce the \emph{scaled dot-product attention}, which is the core component of the Transformer.

The input of the scaled dot-product attention consists of a query $q\in\mathbb{R}^{d}$, a set of keys $k_t\in\mathbb{R}^{d}$ and values $v_t\in\mathbb{R}^{d}$, where $t\in\{1,2,...,n\}$ is the number of key-value pairs and $d$ is the common dimensionality of all the inputs features. We calculate the dot products of query with all keys, divide each by $\sqrt{d}$ and apply a softmax function to obtain the attention weights on the values. In practice, we pack all the keys and values into matrices $K=[k_1,...,k_n]\in\mathbb{R}^{n \times d}$ and $V=[v_1,...,v_n]\in\mathbb{R}^{n \times d}$ respectively. The attention function on a set of queries $Q=[q_1,...,q_m]\in\mathbb{R}^{m\times d} $ can be computed in parallel as follows:
\begin{equation}\label{eq:scaled dot-product attention}
F = A(Q,K,V) = \mathrm{softmax}(\frac{QK^T}{\sqrt{d}})V
\end{equation}
where $F\in\mathbb{R}^{m \times d}$ correspond to the attended features of the queries $Q$.

Instead of performing a single attention function for the queries, multi-head attention is introduced in \cite{vaswani2017attention} to allow the model to attend to diverse information from different representation subspaces. The multi-head attention contains $h$ parallel `heads' with each head corresponding to an independent scaled dot-product attention function. The attended features $F$ of the multi-head attention functions is given as follows:
\begin{equation}\label{eq: multi-head attention}
F = MA(Q,K,V) = \mathrm{Concat}(h_1, ..., h_h)W^o
\end{equation}
\begin{equation}\label{eq: head_j}
h_i = A(QW_i^Q,KW_i^K,VW_i^V)
\end{equation}
where $W_i^Q,W_i^K,W_i^V\in\mathbb{R}^{d \times d_h}$ are the projection matrices of the $i$-th head. $W^O\in\mathbb{R}^{h*d_h\times d}$ is the output projection matrix that aggregates the information from different heads. $d_h$ is the dimensionality of the output features of each head. To prevent the model from becoming too large, we set $d_h = d/h$.

\begin{figure}
\begin{center}
\includegraphics[width=0.49\textwidth]{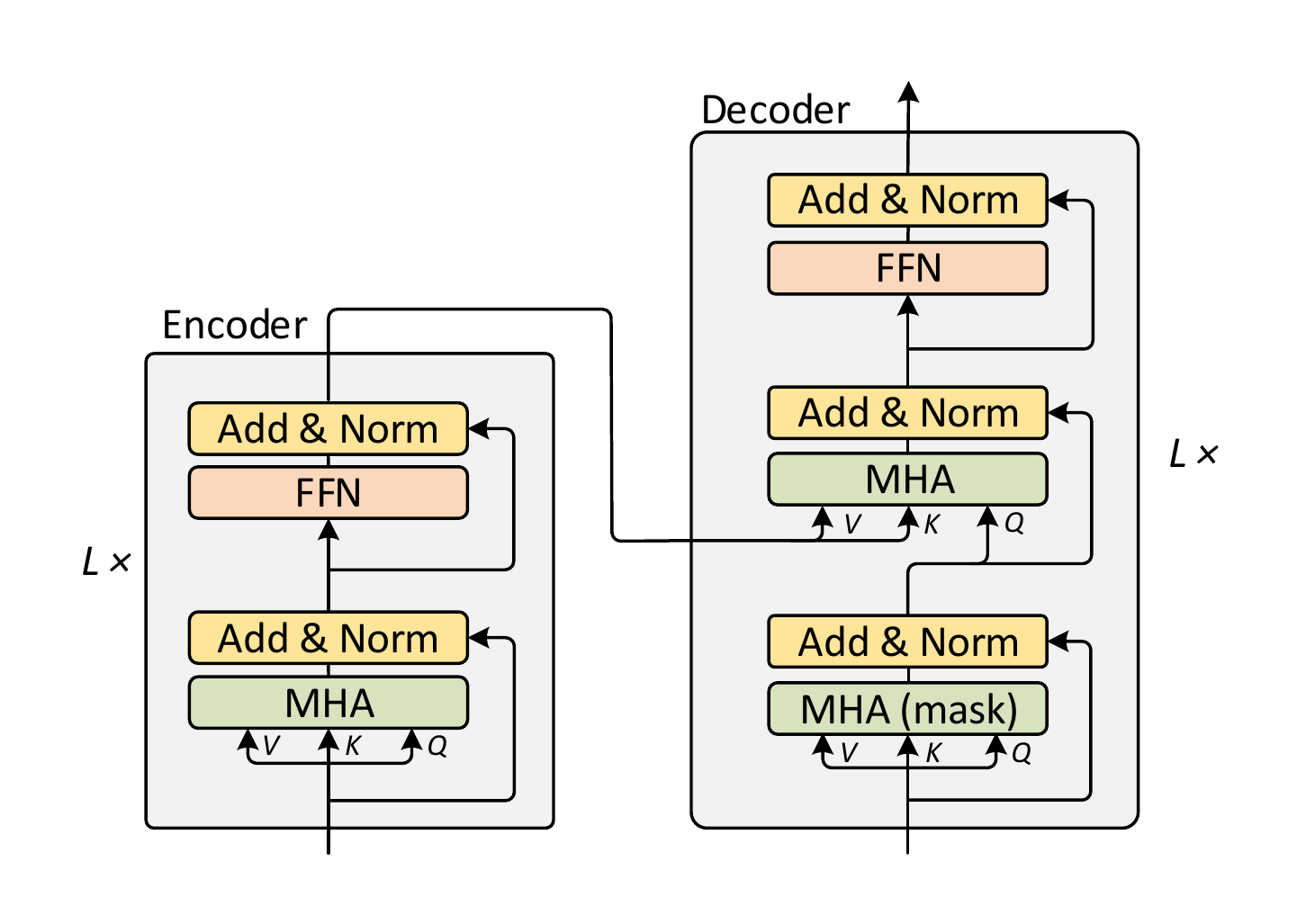}
\caption{Transformer architecture in an encoder-decoder manner. MHA and FFN denote the multi-head attention module and the feed-forward networks module, respectively. $L$ is the number of stacked attention blocks for the encoder and decoder, and is set to the same number for simplicity.}
\label{fig:transformer}
\end{center}
\vspace{-10pt}
\end{figure}

In addition to the multi-head attention (MHA), another basic component in the Transformer is the feed-forward networks (FFN). FFN takes the input from MHA and further transform it using two linear layers with the ReLU activation and dropout \cite{srivastava2014dropout} in between as follows:
\begin{equation}\label{eq:ffn}
\mathrm{FFN}(x) = \mathrm{FC}(\mathrm{Dropout}(\mathrm{ReLU}(\mathrm{FC}(x))))
\end{equation}

The Transformer is a deep end-to-end architecture that stacks attention blocks to form an encoder-decoder strategy (see Fig. \ref{fig:transformer}).
Both the encoder and the decoder consist of $N$ attention block, and each attention block contains the MHA and FFN modules. The MHA module learns the attended features that consider the pairwise interactions between two input features, and the FFN module further nonlinearly transforms the attended features. In the encoder, each attention block is \emph{self-attentional} such that the queries, keys and values in Eq.(\ref{eq:scaled dot-product attention}) refer to the same input features. In contrast, the attention block in the decoder contains a self-attention layer and a guided-attention layer. It first models the self-attention of given input features and then takes the output features of the last encoder attention block to guide the attention learning. To simplify the optimization, shortcut connection \cite{he2016deep} and layer normalization \cite{ba2016layer} are applied after all the MHA and FFN modules.

\subsection{Multimodal Transformer for Image Captioning}\label{sec:mtic}
\begin{figure}
\begin{center}
\includegraphics[width=0.4\textwidth]{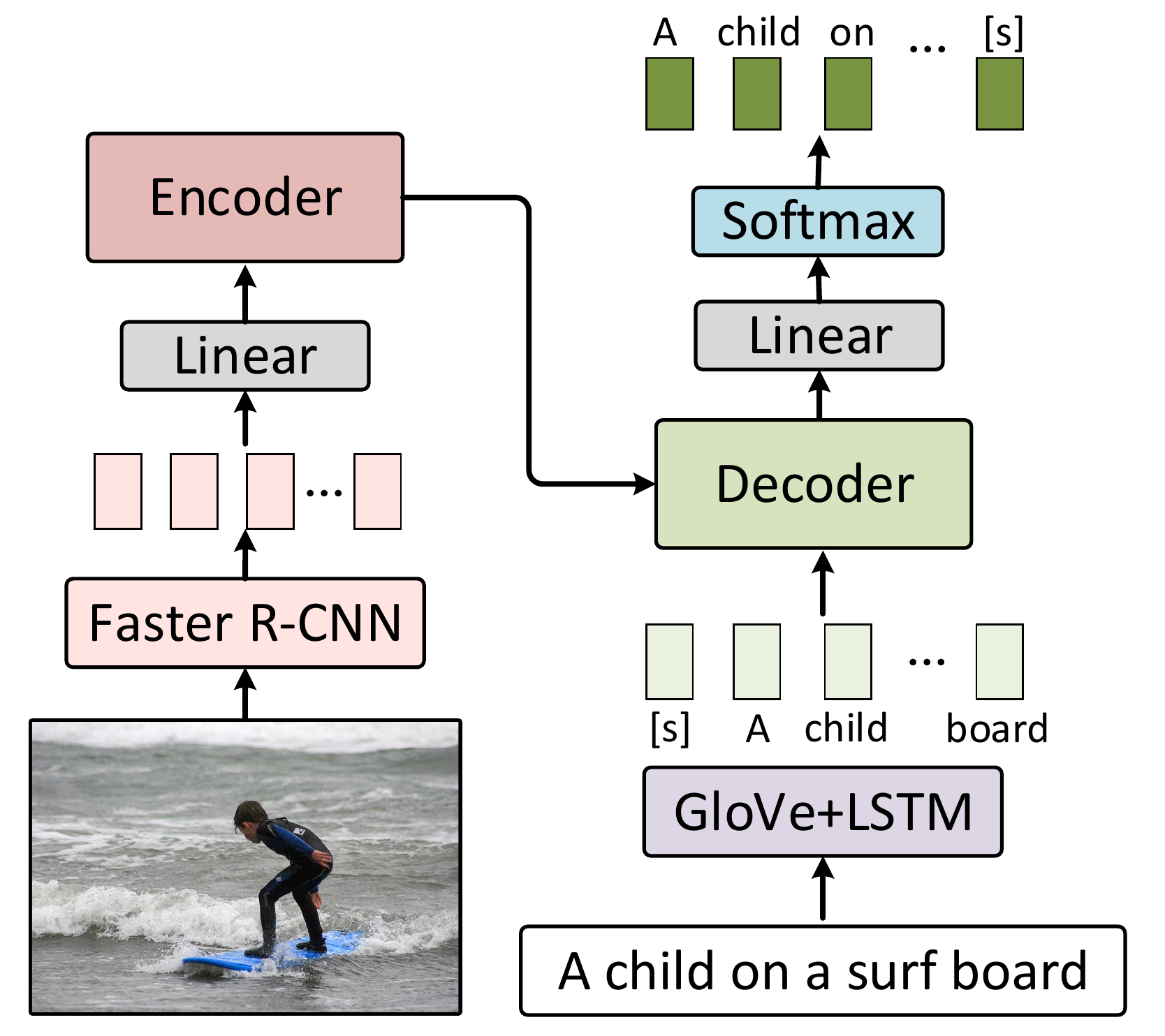}
\caption{Multimodal Transformer (MT) model for image captioning. It consists of an image encoder to learn self-attended visual features, and a caption decoder to generate the caption from the attended visual features. $[\mathsf{s}]$ is a delimiter that indicates the start or the end of the caption.}
\label{fig:mt}
\end{center}
\vspace{-10pt}
\end{figure}

Based on the preliminary information about the Transformer above, we describe the Multimodal Transformer (MT) architecture for image captioning, which consists of an image encoder and a textual decoder. The image encoder takes an image as its input and uses a pre-trained Faster-RCNN model \cite{ren2015faster} to extract region-based visual features. The visual features are then fed into the encoder to obtain the attended visual representation with self-attention learning. The decoder takes the attended visual features and the previous word to predict the next word recursively. The flowchart of the MT architecture is shown in Fig. \ref{fig:mt}.

\noindent \textbf{Image Encoder.} The input image is represented as a group of visual features that are extracted from a pre-trained object detector \cite{anderson2018bottom}. Specifically, the detector is a Faster-RCNN model \cite{ren2015faster} that is pre-trained on the Visual Genome dataset \cite{krishna2017visual}. We sort the detected objects w.r.t. their confidence scores in descending order and keep the top-$m$ objects. Each object is represented as a feature vector $x_i\in\mathbb{R}^{d_x}$ by mean-pooling the convolutional feature from its detected region. Finally, the image is represented as a feature matrix $X\in\mathbb{R}^{m\times d_x}$.

The visual features $X$ is first fed into a fully-connected layer to adapt the feature dimensionality to the encoder. The projected features (denote as $X^{(0)}$) are then fed into the encoder with $L$ attention blocks $[A_\mathrm{enc}^1, A_\mathrm{enc}^2,...,A_\mathrm{enc}^L]$. The $i_{th}$ attention block $A_\mathrm{enc}^l$ takes the output features $X^{l-1}$ from the $i-1_{th}$ attention block, and output their attended features $X^l$ in a recursive manner.
\begin{equation}\label{eq:encoder}
X^l = A_\mathrm{enc}^l(X^{l-1})
\end{equation}
where each $A_\mathrm{enc}^l$ consists of a MHA module and a FFN module with independent model weights (see Fig. \ref{fig:transformer}).

\noindent \textbf{Caption Decoder.} Based on the visual representations from the encoder, the textual decoder generates captions for the image. The input caption is first tokenized into words and trimmed to a maximum length of $n$ words. Each word in the caption is first represented as a word vector $y_i \in \mathbb{R}^{300} $ by using the 300-D GloVe word embedding \cite{pennington2014glove} pre-trained on a large-scale corpus. We use a feature matrix $Y \in \mathbb{R}^{n \times 300}$ to represent a caption sentence. For the captions that are shorter than 16 words, we use zero-padding to fill them to the maximum size. To model the temporal information of the captions, the word embeddings are then pass through a one-layer LSTM network \cite{hochreiter1997long} with $d_y$ hidden units, resulting in caption representations $Y=[y_1,...,y_n]\in\mathbb{R}^{n \times d_y}$.

In the training stage, the caption decoder takes the inputs from both the image encoder and caption representations. Given the attended image features $X^L$ and the caption input features $Y$, the caption decoder with $L$ attention blocks ($[A_\mathrm{dec}^1, A_\mathrm{dec}^2,...,A_\mathrm{dec}^L]$) learns to predict the attended word features in an analogous manner to the strategy in the encoder.
\begin{equation}\label{eq:decoder}
Y^l = A_\mathrm{dec}^l(X^L, Y^{l-1})
\end{equation}
where each $A_\mathrm{dec}^l$ consists of two MHA modules and one FFN module (see Fig. \ref{fig:transformer}). The first MHA module models the self-attentions on the caption words and the second MHA module learns the image-guided attention on the caption words. Note that the self-attention (\emph{i.e.}, the first MHA module) is only allowed to attend to earlier positions in the output sequence and is implemented by masking subsequent positions (setting them to -$\infty$) before the softmax step in the self-attention calculation, thereby resulting in a triangular mask matrix $M\in\mathbb{R}^{n\times n}$. The output features $Y^L=[y^L_1,y^L_1,...,y^L_n]$ are fed into a linear word embedding layer to transform the features to a $d_v$-dimensional space, where $d_v$ is the vocabulary size. Subquently, softmax cross-entropy loss is performed on each word to predict the probability of its next word.

In the testing stage, the caption is generated word-by-word in a sequential manner. When generating the $t_{th}$ word, the input features are represented as $Y_{\leq t}=[y_1,y_2,...y_{t-1},\textbf{0},...,\textbf{0}]\in\mathbb{R}^{n \times d_y}$, where $\textbf{0}\in\mathbb{R}^{d_y}$ corresponds a zero-padded feature. The input caption features along with the image features are fed forward the model to obtain the word with the largest probability among the whole word vocabulary. The predicted word is then integrated into the inputs to recursively generate the new inputs $Y_{\leq t+1}$. To improve the diversity of generated captions, we also introduce the beam search strategy during the testing stage.

\section{Image Encoder with Multi-view Visual Representation}\label{sec:multiview}
\begin{figure*}
\begin{center}
\includegraphics[width=0.97\textwidth]{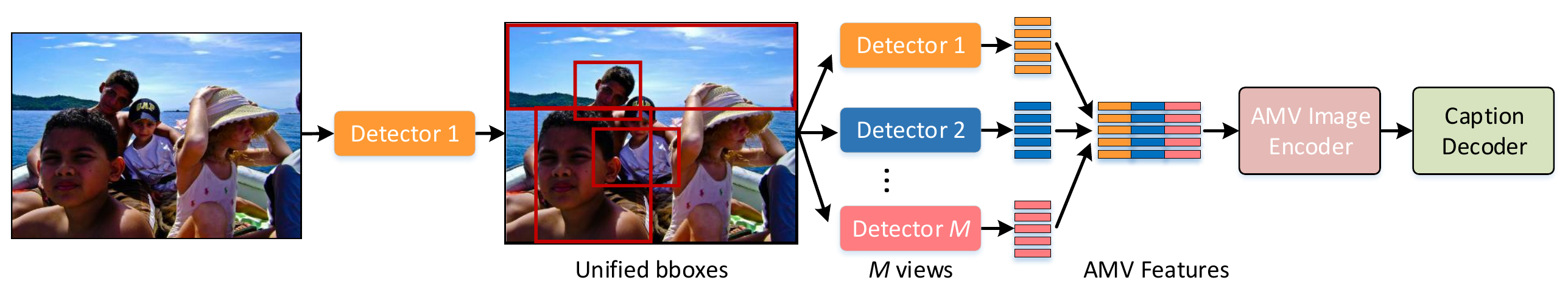}
\caption{The flowchart of the aligned multi-view (AMV) image encoder model. Given an image, different object detectors are regarded as the multiple views. To obtain the aligned multi-view features, we choose one of the $M$ detectors to predict the unified bounding boxes for objects, and then use these bboxes to extract aligned multi-view features. The aligned multi-view features are fed into the AMV image encoder (which is exactly the same as the one for single-view features introduced in section \ref{sec:mtic}).}
\label{fig:amve}
\end{center}
\vspace{-10pt}
\end{figure*}
In this section, we introduce multi-view image representations and modify the the image encoder in section \ref{sec:mtic} to multi-view image encoder to facilitate the representation capacity of the MT model. Though it has been intensively investigated by previous works \cite{yu2014high}\cite{tu2018semantic},\cite{tao2018deep}, existing multi-view learning approaches focus on integrating the \emph{global} multi-view features (\emph{e.g.}, color histogram or GIST descriptor) from the whole image. The global multi-view features may fail to preserve the fine-grained semantics of the image, thus leading to incorrect caption. In contrast, we extract region-based \emph{local} multi-view features to represent the image. Each object detector (\emph{i.e.}, pre-trained Faster R-CNN models with different backbones) is regarded as one single view.

Note that the objects extracted from different detectors are naturally unaligned, thereby making it challenging to learn the correspondence across different views. To address this problem, we extend the proposed image encoder model in section \ref{sec:mtic}, and introduce two multi-view image encoder models, namely, the \emph{Aligned Multi-View} (AMV) image encoder and the \emph{Unaligned Multi-View} (UMV) image encoder, respectively.

\begin{figure*}
\begin{center}
\includegraphics[width=0.99\textwidth]{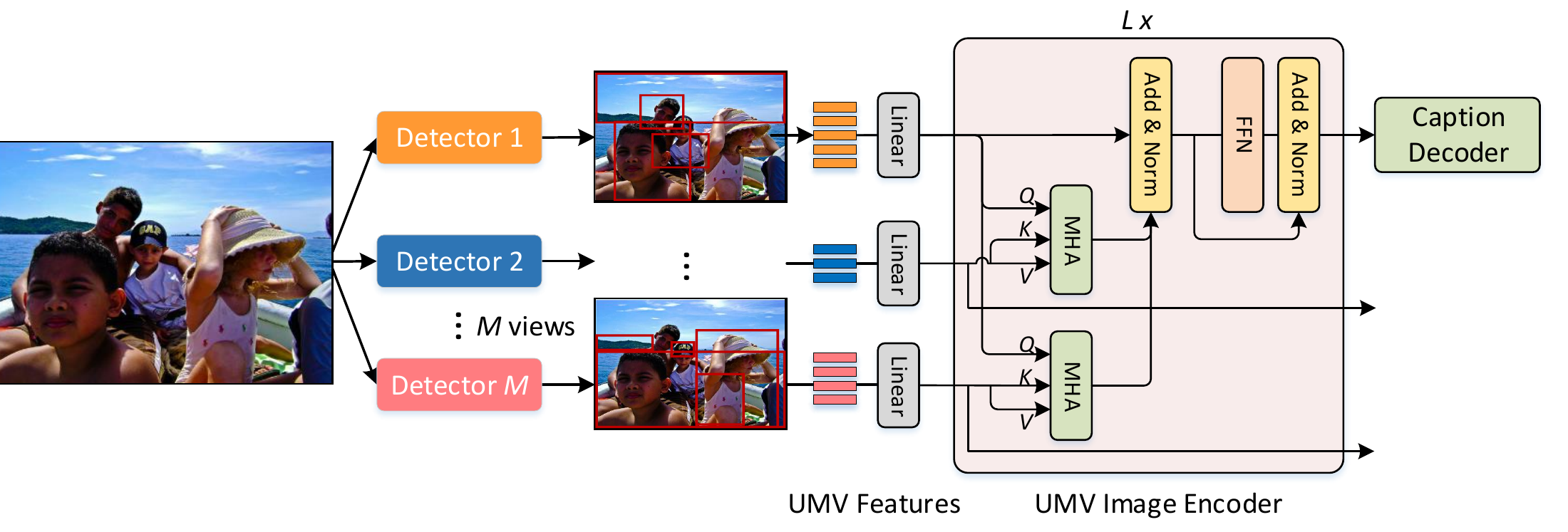}
\caption{The flowchart of the unaligned multi-view (UMV) image encoder model. Given an image, unaligned multi-view features are extracted from different object detectors in parallel. The unaligned multi-view features are fed into the UMV model to output the attended features with adaptive alignment learning.}
\label{fig:umve}
\end{center}
\vspace{-10pt}
\end{figure*}

\subsection{Aligned Multi-View Image Encoder}
The AMV model uses a simple strategy to obtain the aligned multi-view features from different object detectors. Rather than extracting the object bounding boxes and corresponding features for each view, we propose a two-stage feature extraction framework. Given $M$ pre-trained Faster R-CNN models, we first select one detector as the primary model to generate the unified bounding boxes for all views. The choices of different primary models has little influence on the quality of the generated features, and we simply choose the model with the highest detection performance.  Subquently, the unified bounding boxes are used to extract features from different Faster R-CNN models. Specifically, the Faster R-CNN models degenerate to their Fast R-CNN versions \cite{girshick2015fast} that take the pre-computed bounding boxes as inputs. The resulting multi-view features are aligned such that each paired multi-view features correspond to one object in the image.

Assuming that we generate $m$ unified bounding boxes, the extracted features from the $i$-th view ($i\in\{1,2,...,M\}$) can be represented as $X_{(i)}\in\mathbb{R}^{m \times d_i}$, where $d_i$ is the dimensionality of the features. By simply concatenating the features in columns, we obtain the multi-view features $X=[X_{(1)},X_{(2)},...,X_{(M)}]\in\mathbb{R}^{m\times (d_1+d_2+...,d_M)}$. These aligned multi-view features can replace the aforementioned single-view feature, and be seamlessly fed into the image encoder. The overall flowchart of the AMV model is shown in Fig. \ref{fig:amve}.

To align the multi-view features, the AMV model uses the unified bounding boxes. However, we argue that this strategy may harm the diversity of multi-view features, leading to a limited representation capacity of the encoded image features. Moreover, the AMV model implicitly constrains the object detector for each view to be a Faster R-CNN model, which can either take the pre-computed proposals as inputs or generate the proposals using the built-in Region Proposal Networks (RPN) \cite{ren2015faster}. This constraint limits the usage of one-stage object detectors, \emph{e.g.}, RetinaNet \cite{lin2017focal} and YOLO \cite{redmon2016you}.

\subsection{Unaligned Multi-View Image Encoder}

To address the limitations of the AMV encoder model, we propose a more generalized unaligned multi-view (UMV) image encoder model that can directly integrate the unaligned multi-view features from different object detectors (see the flowchart in Fig. \ref{fig:umve}).

The extracted visual features for the $i$-th view can be represented as $X_{(i)}\in\mathbb{R}^{m_i \times d_i}$, where the number of features $m_i$ and the feature dimensionality $d_i$ can be different across multiple views. The unaligned multi-view features are fed into an encoder to be aligned and fused simultaneously. Specifically, we choose one view as the primary view and use its features to guide the attention learning for other views. The attended features from other views are then integrated into the features in the primary view to output the output features.

Given the multi-view features $X_{(1)}, X_{(2)},...,X_{(M)}$, they are first linearly projected into a common $d$-dimensional space to obtain their transformed representations $F_{(1)}, F_{(2)},...,F_{(M)}$.
Assuming that $F_{(1)}$ corresponds to the features of the primary view, we have $M-1$ MHA modules in total to model the interactions between $F_{(1)}$ and $F_{(i)}$ with $i\in\{2,3,...,M\}$.
\begin{equation}\label{eq:umv_mha}
\tilde{F}_{(i)} = \mathrm{MHA}_{(i)}(F_{(1)}, F_{(i)},F_{(i)})
\end{equation}
where $\tilde{F}_{(i)}\in\mathbb{R}^{m_1 \times d}$ is the attended output features for the $i$-th view. The obtained features $\tilde{F}_{(2)}, \tilde{F}_{(3)},...,\tilde{F}_{(M)}$ have the same shape as $F_{(1)}$, and so they can be integrated with $F_{(1)}$ via an element-wise summation. The MHA modules here can be understood as learning the image-guided attention over the image features from other views.
\begin{equation}\label{eq:umv_addnorm}
\tilde{F}_{(1)} = F_{(1)} + \tilde{F}_{(2)}+ \tilde{F}_{(3)},...,+\tilde{F}_{(M)}
\end{equation}
Following the image encoder model in section \ref{sec:mtic}, the integrated features $\tilde{F}_{(1)}$ that are followed by layer normalization \cite{ba2016layer} are then fed forward through the FFN module to obtain the transformed representations. It is worth noting that the UMV model can also be stacked in depth to learn more accurate interactions across different views, thus resulting in more discriminative output visual features for generating captions.

\section{Experiments}\label{sec:exp}
In this section, we conduct experiments and evaluate the proposed MT models on MSCOCO 2015 image captioning dataset \cite{lin2014microsoft}. Additionally, we use the Visual Genome dataset \cite{krishna2017visual} to pre-train the object detectors that are further used to extract the bottom-up-attention visual features \cite{anderson2018bottom}.

\subsection{Datasets}
\noindent \textbf{\textit{MSCOCO}} is a benchmark dataset for various computer vision tasks, including object detection, instance segmentation, and image captioning \cite{lin2014microsoft}. It contains 83k training images, 40k validation images, and 81k test images. Each image is associated with five captions. Similar to \cite{anderson2018bottom}, we use the \emph{Karpathy} splits \cite{karpathy2015deep} that have been extensively used for reporting results in prior works. These splits merge the images from the original train and val splits, resulting in 121k images in total. After that, the 123k images are split into 113k/5k/5k images for training/validation/testing, respectively. The trained models are ensembled to obtain the predictions that are submitted to the official MSCOCO test server. To evaluate the caption quality, we use four automatic evaluation metrics, namely, BLEU \cite{papineni2002bleu}, ROUGE-L \cite{lin2004rouge}, METEOR \cite{banerjee2005meteor} and CIDEr \cite{vedantam2015cider}.

\noindent \textbf{\textit{Visual Genome}} is a large-scale dataset to evaluate the interactions between objects in the images. It contains 108k images with densely annotated objects, attributes, and relationships. Following the strategies in \cite{anderson2018bottom}, we use the object and attribute annotations to pre-train the bottom-up-attention models. All the images are split into training (98k images), validation (5k images) and testing (5k images). Since part of images in Visual Genome are also found in the MSCOCO captioning dataset, we perform careful checking to avoid affecting the MSCOCO validation and testing splits. Similar to \cite{anderson2018bottom}, we perform extensive cleaning and filtering of the training data to select 1,600 object classes and 400 attributes. This cleaned dataset is used for training our object detection models.

\captionsetup[subtable]{font=small}
\begin{table*}
    \scriptsize
	\begin{subtable}[t]{.54\textwidth}
		\centering
        \subcaption{\textbf{Caption Representations:} Scores of the MT$_\mathrm{sv}$ models (ResNet-101 backbone) with different caption representations. The reference model uses randomly initialized word embeddings and then fine-tuned. PE denotes the positional encoding to model the temporal information of the caption \cite{vaswani2017attention}. GloVe$_{\mathrm{pt}}$ and GloVe$_{\mathrm{pt+ft}}$ mean the word embeddings are pre-trained with GloVe, while GloVe$_{\mathrm{pt+ft}}$ is additionally fine-tuned along with the model. }
		\begin{tabular}{lcccccc}
            \toprule
            \multirow{3}{*}{\textbf{Model}} & \multicolumn{3}{c}{\textbf{Cross-Entropy Loss}} & \multicolumn{3}{c}{\textbf{Self-Critical Loss}} \\
            \cmidrule(r){2-4} \cmidrule(r){5-7}
            & B@1 & M & C & B@1 & M & C \\
            \midrule
            Rand$_{\mathrm{ft}}$~+~PE&  76.0 & 28.2 & 115.9 & 80.4 & 28.9 & 129.2\\
            GloVe$_{\mathrm{pt}}$~+~PE &  76.2 & 28.0 & 116.6 & 80.5 & 29.0 & 129.3\\
            GloVe$_{\mathrm{pt}}$~+~LSTM & \textbf{76.2} & \textbf{28.3} & \textbf{117.1} & 80.8 & \textbf{29.1} & 130.8\\
            GloVe$_{\mathrm{pt+ft}}$~+~LSTM & \textbf{76.2} & \textbf{28.3} & \textbf{117.1} & \textbf{81.2} & \textbf{29.1} & \textbf{130.9}\\
            \bottomrule
        \end{tabular}
		\label{table:caption_feat}
	\end{subtable}
    \quad
	\begin{subtable}[t]{.44\textwidth}
		\centering
        \subcaption{\textbf{Number of Attention Blocks:} Scores of the MT$_\mathrm{sv}$ models with different number of attention blocks $L \in \{1, 2, 4, 6,8\}$. For each model, we also report its corresponding number of model parameters.}
		\begin{tabular}{lccccccc}
            \\
            \toprule
            \multirow{3}{*}{\textbf{\textit{L}}} & \multirow{3}{*}{\makecell{\textbf{\#Params}\\($\times 10^6$)}} & \multicolumn{3}{c}{\textbf{Cross-Entropy Loss}} & \multicolumn{3}{c}{\textbf{Self-Critical Loss}} \\
            \cmidrule(r){3-5} \cmidrule(r){6-8}
            & & B@1 & M & C & B@1 & M & C \\
            \midrule
            1 &17.1& 76.3 & 27.9 & 113.7 & 79.4 & 28.3 & 124.3\\
            2 &25.1& 76.4 & 28.3 & 116.6 & 80.1 & 28.6 & 127.2\\
            4 &39.8& \textbf{76.5} & \textbf{28.4} & \textbf{117.1} & 80.4 & 29.0 & 129.6\\
            6 &54.5& 76.2 & 28.3 & 117.0 & \textbf{80.8} & \textbf{29.1} & \textbf{130.9}\\
            8 &69.2& 76.4 & 28.1 & 116.5 & 80.7 & 29.0& 130.4 \\
            \bottomrule
        \end{tabular}
		\label{table:depth}
	\end{subtable}
    \\
    \\
    \\
    \begin{subtable}[t]{.52\textwidth}
	\centering
    \subcaption{\textbf{Single-view \textit{vs.} Multi-view:} Scores of the MT models with single-view feature (MT$_\mathrm{sv}$), aligned multi-view features (MT$_\mathrm{amv}$) or unaligned multi-view features (MT$_\mathrm{umv}$).}
        \begin{tabular}{lccccccc}
            \toprule
            \multirow{3}{*}{\textbf{Model}}& \multirow{3}{*}{\textbf{Views}}& \multicolumn{3}{c}{\textbf{Cross-Entropy Loss}} & \multicolumn{3}{c}{\textbf{Self-Critical Loss}} \\
            \cmidrule(r){3-5} \cmidrule(r){6-8}
            & & B@1 & M & C & B@1 & M & C \\
            \midrule
            \multirow{2}{*}{MT$_\mathrm{sv}$} & R-101 & 76.2 & 28.3 & 117.1 & 80.8 & 29.1 & 130.9  \\
             &R-152 &  76.4 & 28.4 & 117.5 & 81.0 & 29.3 & 131.2 \\
            \midrule
            MT$_\mathrm{amv}$ &R-101 and R-152&  77.0 & \textbf{28.6} & 119.4 & 81.2 & 29.4 & 132.7    \\
            MT$_\mathrm{umv}$ &R-101 and R-152& \textbf{77.1} & \textbf{28.6} & \textbf{119.5} & \textbf{81.6} & \textbf{29.5} & \textbf{133.4}     \\
            \bottomrule
        \end{tabular}
        \label{table:NoI}
    \end{subtable}
    \quad
    \begin{subtable}[t]{.45\textwidth}
	\centering
    \subcaption{\textbf{Number of Views:} Scores of the MT$_\mathrm{umv}$ models with different number of views $M\in\{2,3\}$.}
        \begin{tabular}{lccccccc}
            \toprule
            \multirow{3}{*}{\textbf{\textit{M}}}& \multirow{3}{*}{\textbf{Views}}& \multicolumn{3}{c}{\textbf{Cross-Entropy Loss}} & \multicolumn{3}{c}{\textbf{Self-Critical Loss}} \\
            \cmidrule(r){3-5} \cmidrule(r){6-8}
            & & B@1 & M & C & B@1 & M & C \\
            \midrule
            \multirow{3}{*}{2} & R-101 and R-152 &  {77.1} & {28.6} & {119.5} & \textbf{81.6} & {29.5} & {133.4} \\
             &  R-101 and X-101 &    76.7 & 28.4 & 118.4 & 81.4 & 29.4 & 133.0\\
             &  R-152 and X-101 & 76.7 & 28.5 & 118.8 & 81.5 & 29.4 & 133.2  \\
            \midrule
            3 & \makecell{R-101, R-152 \\ and X-101} & \textbf{77.3} &\textbf{28.7} & \textbf{119.6} & \textbf{81.9}& \textbf{29.5} & \textbf{134.1}   \\
            \bottomrule
        \end{tabular}
        \label{table:NoI}
    \end{subtable}
	\caption{Ablations of the proposed MT models evaluated on the MSCOCO Karpathy test split. B@1, M, and C correspond to the BLEU@1, METEOR and CIDEr scores, respectively. For each model, we report the results optimized with either the cross-entropy loss or the self-critical loss \cite{rennie2017self}. R-101, R-152, X-101 denote the object detector with ResNet-101, ResNet-152 and ResNeXt-101 backbones, respectively. All results are obtained with beam search in the testing stage. The best result for each evaluation metric is bolded.}
    \label{table:aba}
\end{table*}

\subsection{Implementation Details}
For the captions, we perform the pre-processing as follows. All the caption sentences are converted to lower case and tokenized into words with white space. The rare words that occur less than 5 times or do not exist in the pre-trained GloVe vocabulary \cite{pennington2014glove} are discarded, resulting in a vocabulary of 9,343 words. Each word in the caption is represented as word embedding vector by looking-up the GloVe word vocabulary. The out-of-vocabulary words are represented as all-zero vectors.

For the images, we use the pre-trained bottom-up-attention models to detect the objects and extract visual features for the detected objects. For multi-view image representation, we trained up to three Faster R-CNN \cite{ren2015faster} models (\emph{i.e.}, number of views $M$=3) with different backbones, namely ResNet-101 \cite{he2016deep}, ResNet-152 \cite{he2016deep} and ResNeXt-101 \cite{xie2017aggregated}, respectively. For each model, we select the top-100 objects with the highest confidence scores to represent the image, where each object is represented as a vector by mean-pooling the last convolutional feature from its detected region.

The hyper-parameters of the MT models that are used in the experiments are listed as follows. The dimensionality of input image features $d_x$, and the input caption features $d_y$ are 2048 and 512, respectively. According to the recommendation in \cite{vaswani2017attention}, the latent dimensionality $d$ in the MHA module is 512, the number of heads $h$ is 8, and the latent dimensionality for each head $d_h = d/h = 64$. The number of attention blocks $L$ in the encoder and decoder ranges in $\in\{1,2, 4, 6,8\}$.

To train the MT models, we use the Adam solver \cite{kingma2014adam} with a batch size of 10. The base learning rate is set to $\mathrm{min}(1te^{-4} ,3e^{-4})$, where $t$ is the current epoch number that starts at 1. After 6 epochs, the learning rate is decayed by 1/2 after every 3 epochs. All models are first trained for 15 epochs using the cross-entropy loss and then are further trained for additional 10 epochs using the self-critical loss to alleviate the exposure bias during cross-entropy optimization \cite{rennie2017self}.

\subsection{Ablation Studies}
We run a number of ablation experiments on MSCOCO image captioning dataset to explore the effectiveness of the single-view MT models (MT$_\mathrm{sv}$) with different hyper-parameters, as well as its multi-view variants with aligned multi-view image encoder MT$_\mathrm{amv}$ and unaligned multi-view image encoder MT$_\mathrm{umv}$. The results shown in Table \ref{table:aba} are discussed in detail below.

\noindent \textbf{Caption Representations:}  Table I(a) summarizes the ablation experiments on different caption representations for MT$_\mathrm{sv}$ with the number of attention blocks $L$=6. Compared with the reference model that uses randomly initialized word embeddings and positional encoding \cite{vaswani2017attention}, we can see that using the word embeddings that are pre-trained by GloVe \cite{pennington2014glove} brings significant improvements. In addition, introducing other tricks such as replacing PE with an LSTM network to model the temporal information, or fine-tuning the GloVe word embeddings along with the MT model can slightly improve the performance further. Note that the GloVe$_{\mathrm{pt}}$+LSTM model and the GloVe$_{\mathrm{pt+ft}}$+LSTM model report the same performance in the cross-entropy loss stage, as the fine-tuning is performed only in the self-critical loss stage. Directly fine-tuning the GloVe embedding from scratch (\emph{i.e.}, from the cross-entropy loss) leads to inferior performance. This result can be explained as the word embeddings being sensitive to the captioning performance, and training from scratch may degenerate their representation capacity.

\begin{table*}
\small
\centering
\caption{\textbf{Single-model} image captioning performance on the MSCOCO Karpathy test split. The methods marked with * denote using the bottom-up-attention visual features from a pre-trained Faster R-CNN model. R, D, I-v3, I-v4 and IR-v2 denotes the ResNet, DenseNet, Inception-v3, Inception-v4 and Inception-ResNet-v2 model, respectively.}
\label{tab:karpathy}
\begin{tabular}{lccccccccccc}
\toprule
\multirow{3}{*}{\textbf{Model}} & \multirow{3}{*}{\textbf{Backbone}}& \multicolumn{5}{c}{\textbf{Cross-Entropy Loss}} & \multicolumn{5}{c}{\textbf{Self-Critical Loss}} \\
\cmidrule(r){3-7} \cmidrule(r){8-12}
&& B@1 & B@4 & M & R & C & B@1 & B@4 & M & R & C \\
\midrule
{SCST} \cite{rennie2017self} &R-101& - &30.0 & 25.9 & 53.4 & 99.4 & - & 34.2 & 26.7 & 55.7 &  114.0   \\
{ADP-ATT} \cite{lu2017knowing} &R-101& 74.2 & 33.2 & 26.6 & - &108.5 & - & - & - & - & -      \\
{LSTM-A} \cite{yao2017boosting} & R-101& 75.4 & 35.2 & 26.9 & 55.8 & 108.8 & 78.6 & 35.5 & 27.3 & 56.8 & 118.3      \\
{Up-Down} \cite{anderson2018bottom}  &R-101*& 77.2 & 36.2 & 27.0 & 56.4 &113.5 & 79.8 & 36.3 & 27.7 & 56.9 & 120.1      \\
RFNet \cite{jiang2018recurrent} & R, D, I-v3, I-v4 and IR-v2 &  76.4 & 35.8 & 27.4 & 56.5 & 112.5 & 79.1& 36.5 & 27.7 & 57.3 & 121.9\\
{GCN-LSTM} \cite{yao2018exploring} &R-101*& \textbf{77.4} &{37.1} &28.1 &57.2 &117.1 & 80.9 & 38.3 & 28.6 & 58.5 & 128.7      \\
\midrule
MT$_\mathrm{sv}$ (ours)  &R-101*&  76.2& 36.6&28.3 & 56.8&117.1  &80.8&39.8 &29.1 & 59.1 & 130.9 \\
MT$_\mathrm{umv}$ (ours) &R-101, R-152 and X-101* & 77.3 &\textbf{37.4} &\textbf{28.7}& \textbf{57.4} &\textbf{119.6} & \textbf{81.9} & \textbf{40.7} & \textbf{29.5} & \textbf{59.7} & \textbf{134.1}       \\
\bottomrule
\end{tabular}
\end{table*}

\begin{table*}
\small
\centering
\caption{\textbf{Real-time leaderboard} of the state-of-the-art solutions on the online MSCOCO test server (April 21st, 2019). The first split shows the published solutions while the second split shows the unpublished ones.}
\label{tab:official}
\begin{tabular}{lcccccccccccccc}
\toprule
\multirow{3}{*}{\textbf{Model}} & \multicolumn{2}{c}{\textbf{B@1}} & \multicolumn{2}{c}{\textbf{B@2}} & \multicolumn{2}{c}{\textbf{B@3}} & \multicolumn{2}{c}{\textbf{B@4}} & \multicolumn{2}{c}{\textbf{M}} & \multicolumn{2}{c}{\textbf{R}} & \multicolumn{2}{c}{\textbf{C}}\\
\cmidrule(r){2-3} \cmidrule(r){4-5} \cmidrule(r){6-7} \cmidrule(r){8-9} \cmidrule(r){10-11} \cmidrule(r){12-13} \cmidrule(r){14-15}
& c5 & c40 & c5 & c40 & c5 & c40  & c5 & c40 & c5 & c40 & c5 & c40 & c5 & c40\\
\midrule
Google NIC \cite{vinyals2015show} & 71.3 &89.5& 54.2 &80.2& 40.7& 69.4& 30.9& 58.7& 25.4& 34.6& 53.0 &68.2& 94.3& 94.6 \\
M-RNN \cite{mao2015deep} & 71.6 & 89.0 & 54.5 & 79.8 & 40.4 & 68.7 & 29.9 & 57.5 & 24.2 & 32.5 & 52.1 & 66.6 & 91.7 & 93.5\\
LRCN \cite{donahue2015long} &  71.8 & 89.5 & 54.8 & 80.4 & 40.9 & 69.5 & 30.6& 58.5 & 24.7&  33.5 & 52.8 & 67.8 & 92.1 & 93.4\\
{ADP-ATT} \cite{lu2017knowing} & 74.8 & 92.0 & 58.4 & 84.5 & 44.4 & 74.4 & 33.6 & 63.7 & 26.4 & 35.9 & 55.0 & 70.5 & 104.2 & 105.9    \\
{LSTM-A} \cite{yao2017boosting} & 78.7 & 93.7 & 62.7 & 86.7 & 47.6 & 76.5 & 35.6 & 65.2 & 27.0 & 35.4 & 56.4 & 70.5 & 116.0 & 118.0   \\
{SCST} \cite{rennie2017self} & 78.1 & 93.7 & 61.9 & 86.0 & 47.0 & 75.9 & 35.2 & 65.5 & 27.0 & 35.5 & 56.3 & 70.7 & 114.7 & 116.7   \\
{Up-Down} \cite{anderson2018bottom} & 80.2 & 95.2 & 64.1 & 88.8 & 49.1 & 79.4 & 36.9 & 68.5 & 27.6 & 36.7 & 57.1 & 72.4 & 117.9 & 120.5   \\
RFNet \cite{jiang2018recurrent} & 80.4 & 95.0 & 64.9 & 89.3 & 50.1& 80.1& 38.0 & 69.2 & 28.2 & 37.2 & 58.2 & 73.1 & 122.9& 125.1 \\
{GCN-LSTM} \cite{yao2018exploring} & - &- & 65.5 & 89.3 & 50.8 & 80.3 & 38.7 & 69.7 & 28.5 & 37.6 & 58.5 & 73.4 & 125.3 & 126.5   \\
\midrule
SRCB-ML-Lab & 81.1 & 95.4 & 66.0 & 89.8 & 51.5 & 81.3 & 39.7 & 71.3 & 28.4 & 37.3 &  58.5 & 73.1 &  125.3 & 126.7 \\
h-p-hl & 80.5 & 95.0 &  65.3 &  89.6 & 50.9 & 81.1 & 39.0 & 70.9 & 28.7 & 38.2 & 58.6 & 74.1 & 125.0 & 127.2 \\
TecentAI.v2 & 81.1 & 95.5 & 65.7 & 90.0 & 50.8 & 80.9 & 38.6 & 70.1 & 28.6 & 37.7 & 58.7 & 73.7 & 125.4 & 127.8  \\
lun & 81.0 & 95.0& 65.8 & 89.6 &  51.4 &  81.3 &  39.4& 71.2 &  29.1 &  38.5  &58.9&  74.5  &126.9 &  129.6 \\
\midrule
MT (ours)  &\textbf{81.7}&\textbf{95.6}&\textbf{66.8}&\textbf{90.5}& \textbf{52.4} &\textbf{82.4} &\textbf{40.4} &\textbf{72.2} &\textbf{29.4} & \textbf{38.9} & \textbf{59.6} & \textbf{75.0} & \textbf{130.0} & \textbf{130.9}      \\
\bottomrule
\end{tabular}
\end{table*}

\noindent \textbf{Number of Attention Blocks:} Table I(b) shows the performance of the MT$_\mathrm{sv}$ models with different number of attention blocks $L\in\{1,2,4,6,8\}$. We can see that the model size grows linearly as $L$ increases. Regarding the performance, we have two observations as follows: 1) as increasing $L$, the model's performance gradually improves and is saturated at a certain number. This can be explained as a deeper model capturing more complex relationships among objects, providing a more accurate understanding of the image contents. In addition, a deeper model has a larger representation capacity and has a larger risk to overfit the training set, and 2) the optimal model is achieved at different $L$ that are trained with different losses, \emph{i.e.}, $L$=4 for the cross-entropy loss and $L$=6 for the self-critical loss. The reinforcement learning-based self-critical loss provides a more diverse exploration of the hypothesis space to avoid overfitting, and thus it can better utilize the potential of large models.

\noindent \textbf{Single-view \textit{vs.} Multi-view:} Next, we compare the MT model with single-view or multi-view features in Table I(c). We use two Faster R-CNN models with different backbones (ResNet-101 or ResNet-152) to extract the multi-view features. For MT$_\mathrm{amv}$, the unified object boxes are extracted from the detector with the ResNet-152 backbone. From the results, we can see following that: 1) the representation capacity of the object detectors may slightly influence the image captioning performance. The MT$_\mathrm{sv}$ model with the ResNet-152 backbone steadily outperforms the counterpart with the ResNet-101 backbone; and 2) introducing multi-view features significantly improves the captioning performance over the single-view models. MT$_\mathrm{umv}$ slightly outperforms MT$_\mathrm{amv}$, thus highlighting the effect of using diverse multi-view features with unaligned objects.

\noindent \textbf{Number of Views:} In Table I(d), we show the performance of the MT$_\mathrm{umv}$ models with different number of views $M$. We can see that when $M=2$, different backbone combinations have little influence on the captioning performance. Moreover, increasing the number of views $M$ from 2 to 3 results in a slight performance improvement for MT$_\mathrm{umv}$, thus indicating that the model is nearly saturated. Therefore, we do not further introduce more views to the image encoder.

\begin{figure*}
\begin{center}
\includegraphics[width=0.97\textwidth]{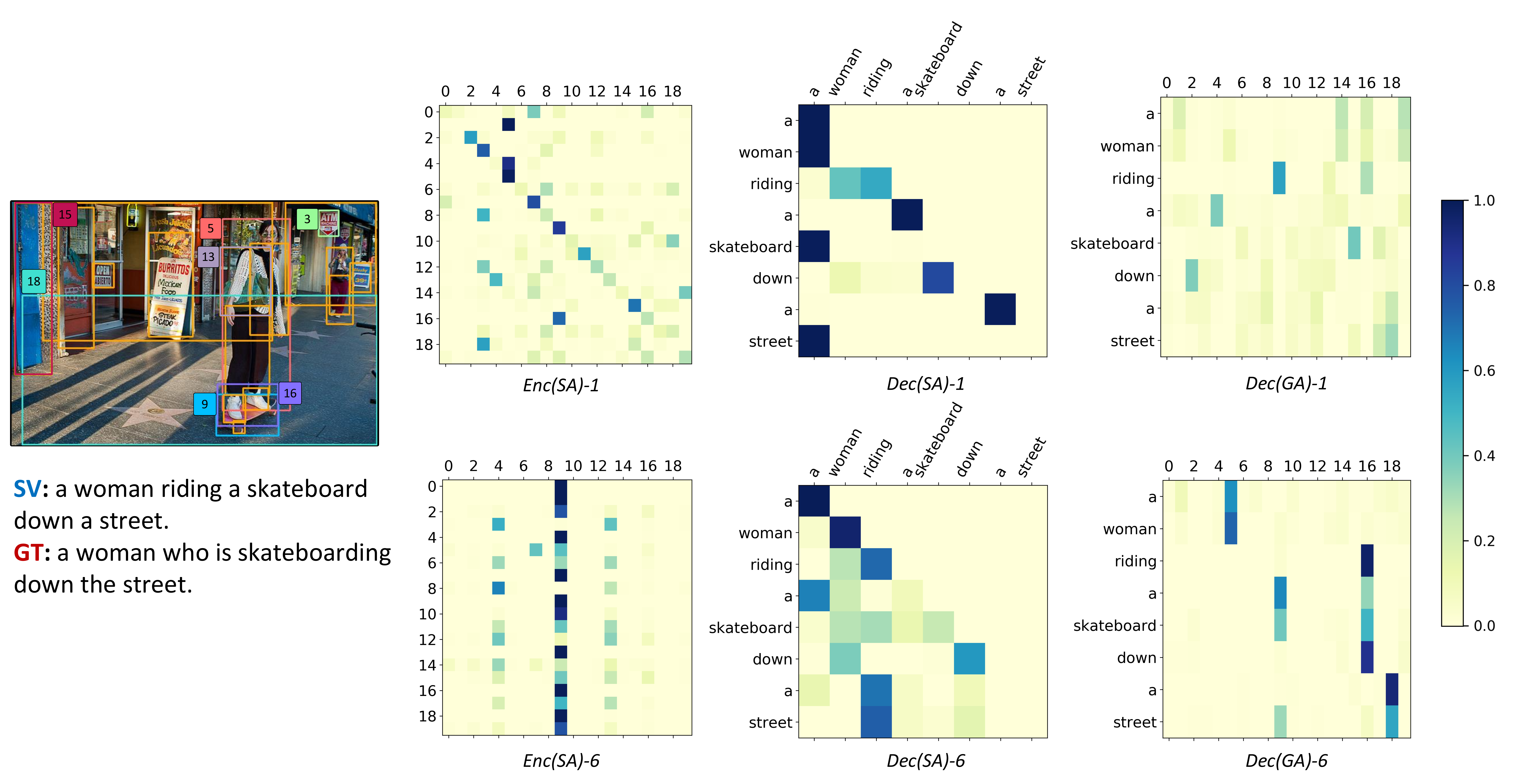}
\end{center}
\caption{Visualizations of the 1st and 6th attention maps ($\mathrm{softmax}(QK/\sqrt{d})$) of the MT$_\mathrm{sv}$ model with R-101 backbone. Enc(SA) denotes the self-attention in the image encoder; Dec(SA) and Dec(GA) denote the self-attention and guided-attention in the caption decoder, respectively. GT denotes the one of the five ground-truth captions provided by MSCOCO. The index within [0-19] shown on the axes of the attention maps corresponds to each object in the image (20 objects in total) . For better visualization effect, we highlight some objects in the image that receive have attention values.}
\label{fig:vis_sv}
\vspace{-10pt}
\end{figure*}

\subsection{Comparison with the State-of-the-Art}
By taking the ablation results into account, we compare our best single-view and multi-view MT models to the current state-of-the-art approaches.

\noindent \textbf{Results on the Karpathy test split:} In Table \ref{tab:karpathy}, we report the comparative results of our approaches along with the SCST \cite{rennie2017self}, ADP-ATT \cite{lu2017knowing}, LSTM-A \cite{yao2017boosting}, Up-Down \cite{anderson2018bottom} and GCN-LSTM \cite{yao2018exploring} on the Karpathy test split. Note that all the compared methods use the same ResNet-101 backbone. With single-view features, the MT$_\mathrm{sv}$ model outperforms most state-of-the-art methods, especially when it is optimized using the self-critical loss. When equipped with multi-view features, the MT$_\mathrm{umv}$ model (trained with the self-critical loss) achieves the new state-of-the-art single-model performance for this split in terms of all evaluation metrics. Note that the RFNet \cite{jiang2018recurrent} also incorporates multi-view features, and they introduce more views than our approach (4 vs. 2). However, its performance is inferior to MT$_\mathrm{umv}$, which suggests that the strategy to fuse multi-view features, rather than the number of views, is the key to the captioning performance.

\noindent \textbf{Results on the official test server:} We also submitted the results of seven MT model ensemble (the MT$_\mathrm{sv}$, MT$_\mathrm{amv}$ and MT$_\mathrm{umv}$ models with different random seeds) to the official MSCOCO test server.\footnote{\url{https://competitions.codalab.org/competitions/3221#results}} Table \ref{tab:official} demonstrates the results of the comparison to the state-of-the-art solutions on the leaderboard including the published ones (in the first split) and the unpublished ones (in the second split). C5 (or c40) denotes the official test settings with 5 (or 40) ground-truth captions, respectively. Compared to all the top performing solutions on the leaderboard, our solution significantly outperforms all the other solutions in terms of all reported evaluation metrics at the time of submission (April 21st, 2019).


\begin{figure*}
\begin{center}
\includegraphics[width=0.99\textwidth]{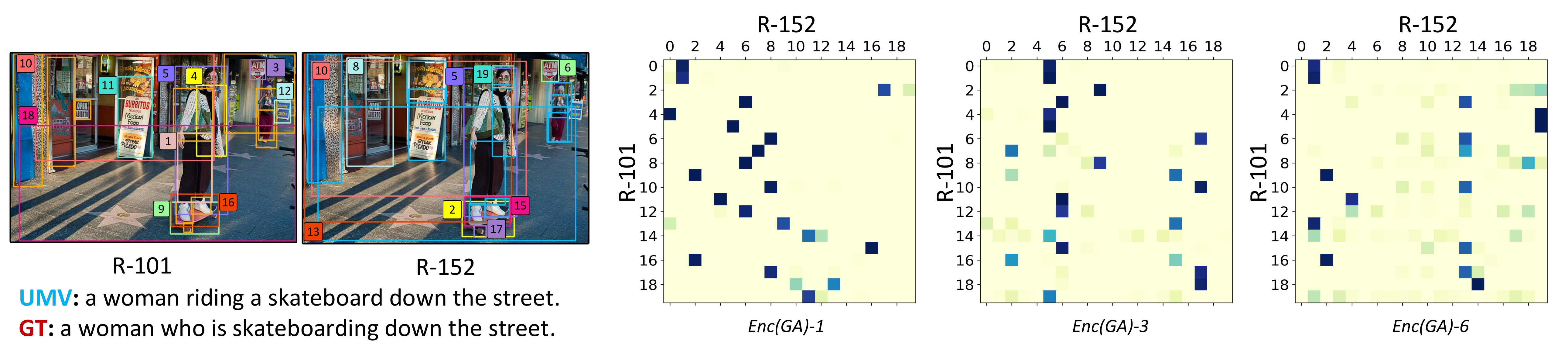}
\end{center}
\caption{Visualizations of the 1st, 3rd and 6th  attention maps of the MT$_\mathrm{umv}$ model with R-101 and R-152 backbones. Enc(GA) denotes the guided-attention in the multi-view image encoder.}
\label{fig:vis_umv}
\vspace{-10pt}
\end{figure*}

\begin{figure*}
\begin{center}
\includegraphics[width=1.0\textwidth]{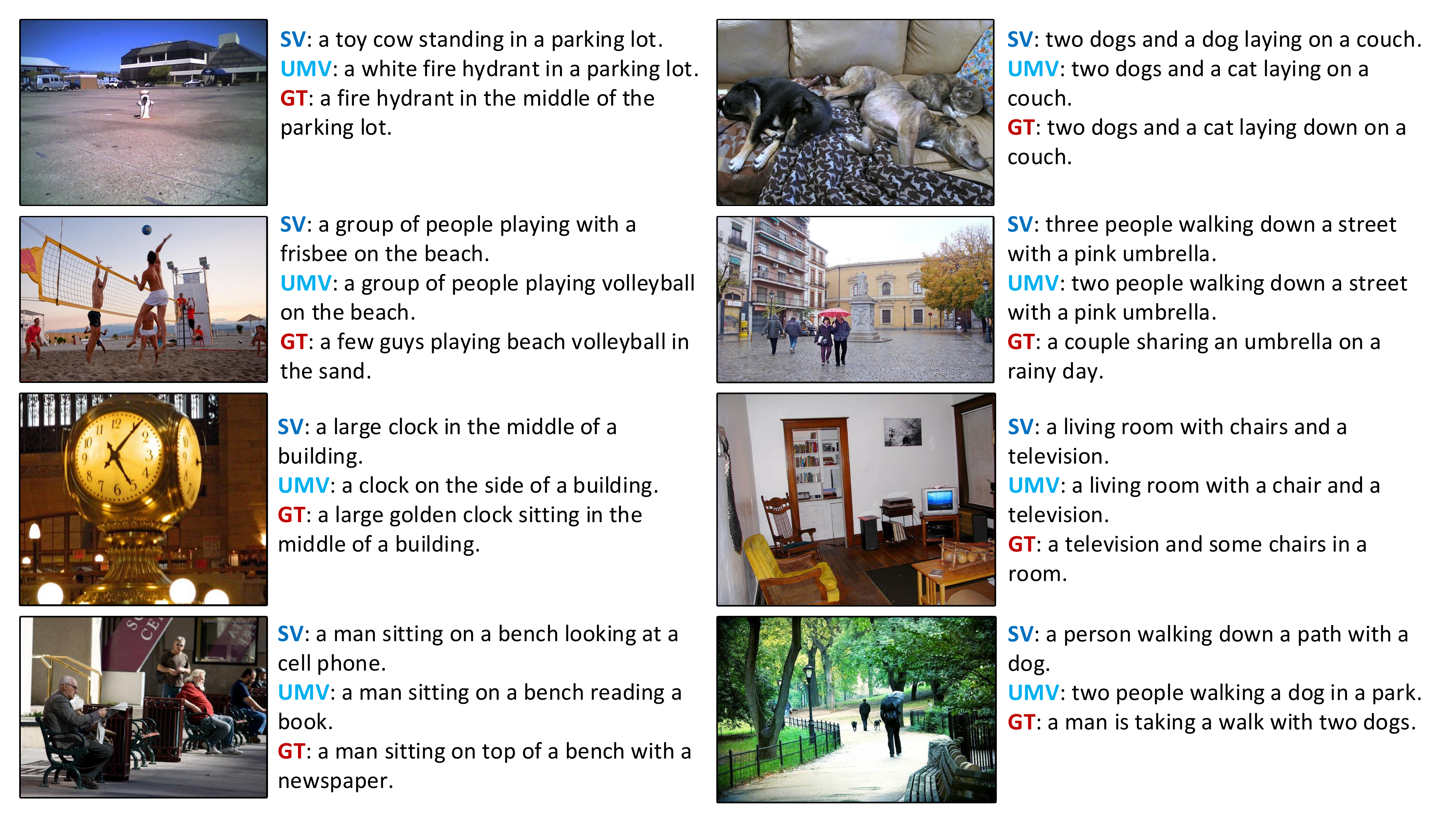}
\end{center}
\caption{Examples generated by the MT$_\mathrm{sv}$ and MT$_\mathrm{umv}$ models on MSCOCO validation set. GT denotes one of the five ground-truth captions. The first two rows show four examples that MT$_\mathrm{umv}$ outperforms MT$_\mathrm{sv}$, and the third row shows two examples that MT$_\mathrm{sv}$ outperforms MT$_\mathrm{umv}$. The last row shows two examples that both models generate incorrect captions.}
\label{fig:vis_examples}
\vspace{-10pt}
\end{figure*}

\subsection{Qualitative Analysis}
To better understand the effectiveness of the proposed approach, we visualize the learned attentions of MT$_\mathrm{sv}$ and MT$_\mathrm{umv}$ in Fig. \ref{fig:vis_sv} and Fig. \ref{fig:vis_umv}, respectively. Due to space limitations, we only show one example for each model and visualize the attention maps from typical attention blocks. From the demonstrated results, we have the following observations.

\noindent \textbf{Attentions of the MT$_\mathrm{sv}$ Encoder:} The self-attentions (SA) of the 1st and 6th blocks in the image encoder that are in Fig. \ref{fig:vis_sv} reflect the pairwise similarity of the visual objects. From the results, we can see that the following: 1) in Enc(SA)-1, the largest attention values almost appear on the diagonal line, indicating that the pairwise interactions are not learned in the first block; and 2) the largest values in Enc(SA)-6 form vertical lines (\emph{e.g.}, the 4th, 9th and 13th columns), which correspond to the key objects of the image (\emph{e.g.}, the girl and the skateboard). This result reveals that all the attended features tend to use the features of these key objects for the representation.

\noindent \textbf{Attentions of the MT$_\mathrm{sv}$ Decoder:} The self-attention the 1st and 6th blocks of the caption decoder that are shown in Fig. \ref{fig:vis_sv} reflects the similarity of paired words. The largest attention values in Dec(SA)-1 almost appear on the diagonal line, which is similar to those in the Enc(SA)-1. In Dec(SA)-6, the word importance and pairwise word similarities are simultaneously represented. For example, the columns of `\emph{woman}' and `\emph{riding}' obtain focused attention weights, and the relationship between `\emph{woman}' and `\emph{skateboard}' is highlighted.

The guided-attention (GA) reflects the multimodal relationships between word-object pairs. In Dec(GA)-1, the learned attentions are not concentrated, and some word-object similarities are incorrect (\emph{e.g.}, the 15th object is not related to the word `\emph{skateboard}'). In contrast, the attention in Dec(GA)-6 has much clearer meanings. The co-attention of key objects along with their word-object relationships are highlighted accordingly.

\noindent \textbf{Attentions of the MT$_\mathrm{umv}$ Encoder:} In Fig \ref{fig:vis_umv}, we visualize the 1st, 3rd and 6th guided-attention (GA) blocks in the multi-view image encoder. In Enc(GA)-1, the unaligned objects from different views are adaptively aligned (\emph{e.g.}, the 5th object in R-101 and the 5-th object in R-152, and the 3rd object in R-101 and the 6th object in R-152). In Enc(GA)-3, the contextual relationships are also involved (\emph{e.g.}, the 5th object in R-152 has large attention values to the 1st and the 4th objects in R-101, which correspond to different parts of the woman's body). In Enc(GA)-6, the modeled contextual relationships cover specific objects and contain background scenes (\emph{e.g.}, the 13th object in R-152 and the 10-th object in R-101). These observations reveal that the UMV image encoder learns to align the objects and explores more complex interactions across multi-view features to provide a fine-grained understanding of the image content.

Moreover, we show some predicted captioning examples in Fig \ref{fig:vis_examples}. The first two rows show four examples where MT$_\mathrm{umv}$ outperforms MT$_\mathrm{sv}$, and the third row shows two examples where MT$_\mathrm{sv}$ outperforms MT$_\mathrm{umv}$. The last row shows two examples where both models generate incorrect captions. From the demonstrated results, we can see the following that: 1) although MT$_\mathrm{umv}$ quantitatively outperforms MT$_\mathrm{sv}$, the performance gap is not qualitatively different and they have their own advantages in different cases. This results in a diverse ensemble when they are integrated together; 2) the incorrect captions are caused by small objects (\emph{e.g.}, the newspaper or the second person).

\section{Conclusions}\label{sec:conclusion}
In this paper, we present a novel Multimodal Transformer (MT) framework for image captioning. The MT consists of an image encoder that generates visual representations via deep self-attention learning, and a caption decoder to transform the encoder's visual features to textual captions. To further facilitate the capacity of visual features, we introduce multi-view learning into the image encoder and propose two MT variants, MT$_\mathrm{amv}$ and MT$_\mathrm{umv}$, to model the aligned multi-view features and unaligned multi-view features, respectively. We quantitatively and qualitatively evaluate the proposed MT models on the benchmark MSCOCO image captioning dataset and conduct extensive ablation studies to explore the reasons behind the MT' s effectiveness. Experimental results show that our method significantly outperforms existing approaches, and an ensemble of seven models achieves the best performance on the real-time leaderboard of the MSCOCO image captioning challenge.
\ifCLASSOPTIONcaptionsoff
  \newpage
\fi



\bibliographystyle{IEEEtran}
\bibliography{TCSVT-03243-2019}

\begin{IEEEbiography}[{\includegraphics[height=1.2in,clip,keepaspectratio]{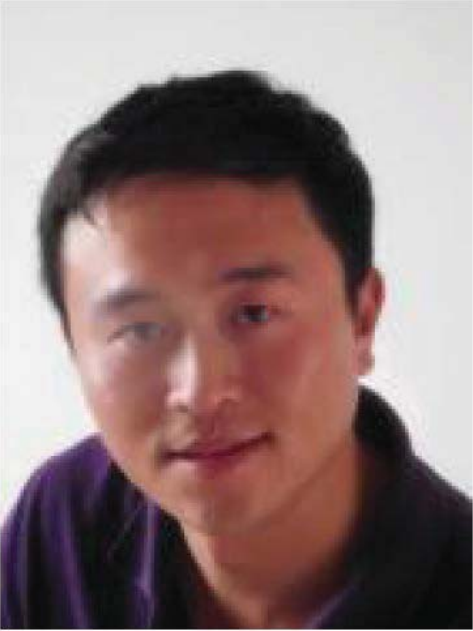}}]{Jun Yu}(M'13) received the B.Eng. and Ph.D. degrees from Zhejiang University, Zhejiang, China. He was an Associate Professor with the School of Information Science and Technology, Xiamen University, Xiamen, China. From 2009 to 2011, he was with Nanyang Technological University, Singapore.
From 2012 to 2013, he was a Visiting Researcher at Microsoft Research Asia (MSRA). He is currently a Professor with the School of Computer Science and Technology, Hangzhou Dianzi University, Hangzhou, China. He has authored or coauthored more than 100 scientific articles. Over the past years, his research interests have included multimedia analysis, machine learning, and image processing. He is the associate editor of IEEE Trans. on CSVT and Pattern Recognition, and the reviewer of various international journals including IEEE Trans. on PAMI, IEEE Trans. on Image Processing, IEEE Trans. on Multimedia, etc. In 2017, he received the IEEE SPS Best Paper Award. Dr. Yu has (co-)chaired
several special sessions, invited sessions, and workshops. He served as a program committee member or reviewer of top conferences and prestigious
journals. He is a Professional Member of the Association for Computing Machinery and the China Computer Federation.
\end{IEEEbiography}
\vspace{-15pt}
\begin{IEEEbiography}[{\includegraphics[height=1.2in,clip,keepaspectratio]{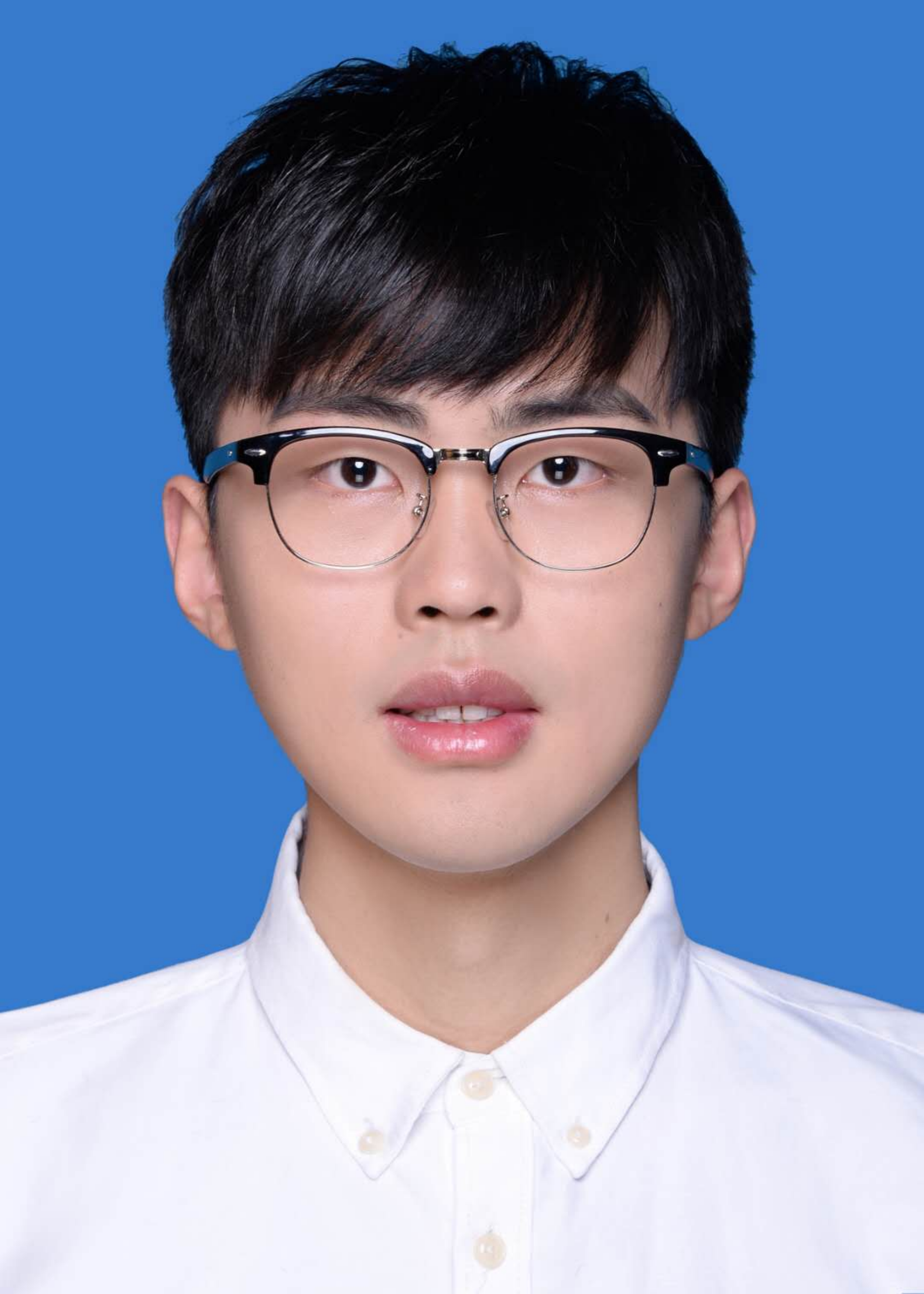}}]{Jing Li}
received the B.Eng. degree from the School of Management, Hangzhou Dianzi University, Hangzhou, China, in 2017. He is currently pursuing the M.Eng. degree with the School of Computer Science and Technology, Hangzhou Dianzi University, Hangzhou, China. His current research interests include multimodal analysis, computer vision and machine learning.
\end{IEEEbiography}
\vspace{-15pt}
\begin{IEEEbiography}[{\includegraphics[height=1.2in,clip,keepaspectratio]{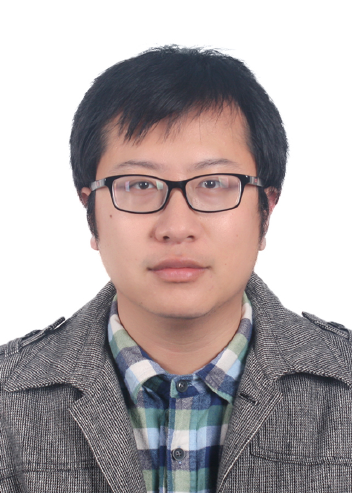}}]{Zhou Yu}
received the B.Eng. and Ph.D. degrees from Zhejiang University, Zhejiang, China, in 2010 and 2015, respectively. He is currently an Associate Professor with the School of Computer Science and Technology, Hangzhou Dianzi University, Hangzhou, China. His research interests includes multimodal analysis, computer vision, machine learning and deep learning. He has served as reviewers or program committee members of prestigious journals and top conferences including IEEE Trans. on CSVT, IEEE Trans. on Multimedia, IEEE Trans. on Image Processing, IJCAI and AAAI, etc.
\end{IEEEbiography}
\vspace{-15pt}
\begin{IEEEbiography}[{\includegraphics[height=1.2in,clip,keepaspectratio]{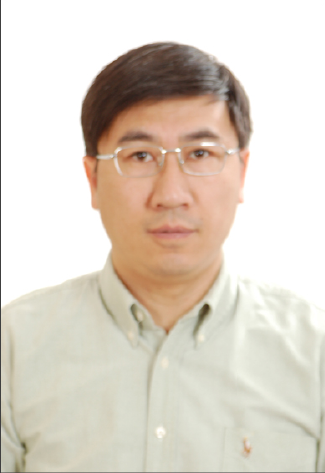}}]{Qingming Huang}(F'18)  is a professor in the University of Chinese Academy of Sciences and an adjunct research professor in the Institute of Computing Technology, Chinese Academy of Sciences. He graduated with a Bachelor degree in Computer Science in 1988 and Ph.D. degree in Computer Engineering in 1994, both from Harbin Institute of Technology, China. His research areas include multimedia computing, image processing, computer vision and pattern recognition. He has authored or coauthored more than 400 academic papers in prestigious international journals and top-level international conferences. He is the associate editor of IEEE Trans. on CSVT and Acta Automatica Sinica, and the reviewer of various international journals including IEEE Trans. on PAMI, IEEE Trans. on Image Processing, IEEE Trans. on Multimedia, etc. He is a Fellow of IEEE and has served as general chair, program chair, track chair and TPC member for various conferences, including ACM Multimedia, CVPR, ICCV, ICME, ICMR, PCM, BigMM, PSIVT, etc.
\end{IEEEbiography}





\end{document}